\documentclass[conference]{IEEEtran}
\IEEEoverridecommandlockouts
\usepackage{cite}
\usepackage{amsmath,amssymb,amsfonts}
\usepackage{algorithmic}
\usepackage{graphicx}
\usepackage{textcomp}
\usepackage{xcolor}

\usepackage{float}
\usepackage{algorithm}
\usepackage{booktabs}
\usepackage[pagebackref,breaklinks,colorlinks]{hyperref}

\def\BibTeX{{\rm B\kern-.05em{\sc i\kern-.025em b}\kern-.08em
    T\kern-.1667em\lower.7ex\hbox{E}\kern-.125emX}}
\begin{document}

\makeatletter % changes the catcode of @ to 11
\newcommand{\linebreakand}{%
  \end{@IEEEauthorhalign}
  \hfill\mbox{}\par
  \mbox{}\hfill\begin{@IEEEauthorhalign}
}
\makeatother % changes the catcode of @ back to 12

\title{A Self-Learning Multimodal Approach for Fake News Detection}
% \author{\IEEEauthorblockN{Anonymous Authors}}

%arxiv
\author{
\IEEEauthorblockN{Hao Chen}
\IEEEauthorblockA{
CUIT\\
%Chengdu, China \\
haochen@cuit.edu.cn}
\and
\IEEEauthorblockN{Hui Guo}
\IEEEauthorblockA{
University at Buffalo\\
% SUNY, NY, USA \\
hguo8@buffalo.edu}
\and
\IEEEauthorblockN{Baochen Hu}
\IEEEauthorblockA{
Dropbox Inc.\\
%San Francisco, USA \\
baochenh@dropbox.com
\and
\IEEEauthorblockN{Shu Hu}
\IEEEauthorblockA{
Purdue University\\
%Indianapolis, USA \\
hu968@purdue.edu}
\and
\linebreakand
\IEEEauthorblockN{Jinrong Hu}
\IEEEauthorblockA{CUIT\\
%Chengdu, China \\
hjr@cuit.edu.cn}
\and
\IEEEauthorblockN{Siwei Lyu}
\IEEEauthorblockA{
University at Buffalo\\
% SUNY, NY, USA \\
siweilyu@buffalo.edu}
\and
\IEEEauthorblockN{Xi Wu}
\IEEEauthorblockA{CUIT\\
%Chengdu, China \\
wuxi@cuit.edu.cn}
\and
\IEEEauthorblockN{Xin Wang}
\IEEEauthorblockA{
University at Albany\\
%SUNY, NY, USA \\
xwang56@albany.edu}
}
}
\maketitle

\begin{abstract}
The rapid growth of social media has resulted in an explosion of online news content, leading to a significant increase in the spread of misleading or false information. While machine learning techniques have been widely applied to detect fake news, the scarcity of labeled datasets remains a critical challenge. Misinformation frequently appears as paired text and images, where a news article or headline is accompanied by a related visuals. In this paper, we introduce a self-learning multimodal model for fake news classification. The model leverages contrastive learning, a robust method for feature extraction that operates without requiring labeled data, and integrates the strengths of Large Language Models (LLMs) to jointly analyze both text and image features. LLMs are excel at this task due to their ability to process diverse linguistic data drawn from extensive training corpora. Our experimental results on a public dataset demonstrate that the proposed model outperforms several state-of-the-art classification approaches, achieving over 85\% accuracy, precision, recall, and F1-score. These findings highlight the model's effectiveness in tackling the challenges of multimodal fake news detection.
\end{abstract}


\section{Introduction}
The emergence of social media platforms has profoundly transformed news dissemination, offering immediate and widespread access to diverse information \cite{wang2022gan}. 
However, this increased accessibility has inadvertently facilitated the rapid spread of misinformation, a problem exacerbated by technologies such as deepfakes \cite{35}. A piece of misinformation is illustrated in Fig.~\ref{fakenews}, where it is evident that images are paired with misleading textual content, which can be easily fabricated or manipulated using AI-driven tools \cite{guo2022eyes}. As a result, such misinformation can rapidly proliferate across the vast expanse of the digital landscape. 
In recent years, the pervasive distribution of false narratives has become a critical societal issue, causing negative impacts both within digital environments and in broader societal contexts. This situation has raised substantial concerns across various demographic groups, leading to heightened anxiety and a decline in public trust toward media sources \cite{pu2022learning}. 
Therefore, there is an urgent need for the development and implementation of effective detection systems to combat the spread of fake news on social media platforms.

\begin{figure}[t]
\centering
\includegraphics[scale=0.6]{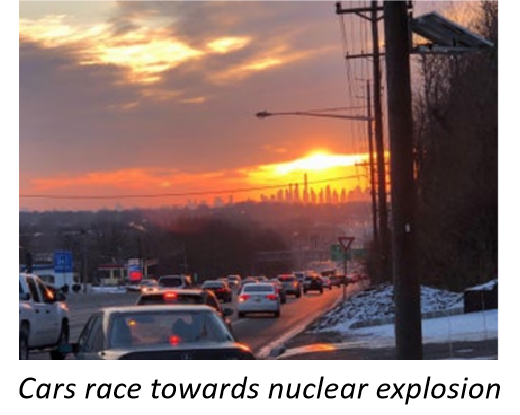}
\caption{An example of fake news (mismatching image-text) from dataset \cite{8}.}
\label{fakenews}
\end{figure}

The labor-intensive and time-consuming nature of manual fact-checking has spurred the development of automated approaches to tackle the widespread issue of fake news. 
Among these, machine learning techniques-particularly supervised classification models \cite{20,27,28}-have garnered significant attention. However, the efficacy of these models largely depends on the availability of high-quality labeled datasets. Unfortunately, such datasets are often difficult to obtain and are typically insufficient in capturing the full diversity inherent in fake news content due to their limited scope. In contrast, the use of weakly supervised or unsupervised methods mitigates the need for large volumes of labeled data and offers distinct advantages over traditional supervised approaches.

While existing approaches primarily focus on textual semantic and syntactic similarities and have achieved some success in fake news detection, they often fail to account for the intricate interactions between different data-modalities, particularly the subtle and complex relationships between images and text. This limitation may undermine the accuracy and robustness of these methods in addressing the issue of misinformation. To address these shortcomings, we propose a novel methodology for multimodal fake news detection. Our approach leverages the advantages of contrastive learning to mitigate the challenge of limited labeled data. Additionally, we incorporate a large language model to integrate and analyze both image and text features, thereby enhancing the model's ability to assess the veracity of news content. Our contributions can be summarized as follows:
\begin{enumerate}
    \item The power of contrastive learning is harnessed to enhance the efficiency of the model's learning in relation to image content. It displays superior detection performance in scenarios with restricted data availability.
    \item The integration of text and image features is accomplished through the utilization of large language models. Specifically, by leveraging learnable queries to align the multimodal features and designing appropriate prompts, the model achieves a heightened accuracy in the detection of fake news.
    \item The dynamic optimization of the loss function enables the model to handle situations where the large language model is fine-tuning. 
\end{enumerate}

\begin{figure*}[t]
\centering
\includegraphics[scale=0.60]{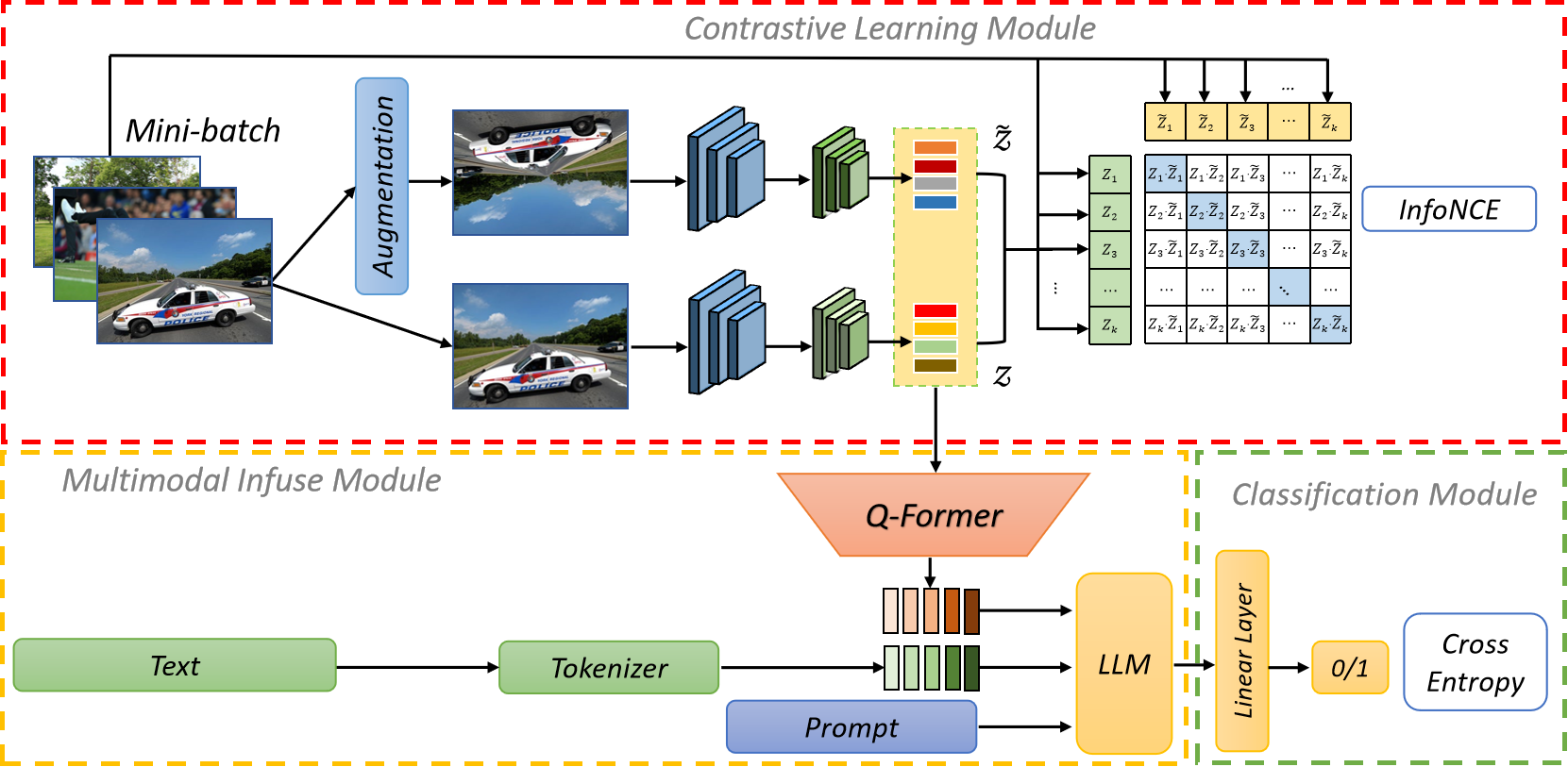}
\caption{The overall structure of multimodal fake news detection. The model is composed of three components, contrastive learning module is for learning the image feature using a small sample of training data, infusing module aims to align text and image feature and then apply the large language model for the multimodal combination, the classification module is for the prediction of fake news.}
\label{framework}
\end{figure*}

It's important to note that most of the techniques we introduce are general and can be applied to various classification tasks. Specifically, our use of contrastive learning proves advantageous in scenarios with a scarcity of labeled training data. The paper is organized as follows: Section~\ref{sec:2} reviews related work, and Section~\ref{sec:3} describes our proposed method. Our experimental evaluation is presented in Section~\ref{sec:4}. The conclusion and future work are presented in Section~\ref{sec:5}.

\section{Related Work}
\label{sec:2}
In recent years, the widespread dissemination of fake news on social media platforms has resulted in significant detrimental effects, thereby motivating the scholarly community to engage in extensive investigations into fake news detection. Initially, Bondielli et al. \cite{9} conducted a rigorous classification of information, distinguishing between fake news and rumors based on whether the content had been thoroughly verified by credible sources. Subsequently, both Meel et al. \cite{11} and Guo et al. \cite{10} provided comprehensive analyses of the various terminologies related to misinformation prevalent on social media, including disinformation, fake news, and misinformation, among others \cite{guo2022open}. Instead of concentrating on the subtle and complex distinctions among these definitions, our research is primarily focused on the machine learning methodologies employed for detection. 
Furthermore, our study predominantly examines news articles that feature paired text and images. Within this framework, the present paper categorizes existing fake news detection techniques into two main types—unimodal and multimodal—based on the nature of the data utilized.

\subsection{Unimodal Classification}
A substantial body of research has leveraged supervised learning algorithms, such as Support Vector Machines (SVM) \cite{20}, Na\"ive Bayes \cite{22}, and Logistic Regression \cite{21}, for the detection of fake news. These models are trained on annotated datasets that classify news articles as either authentic or deceptive. The performance of these algorithms, however, is highly contingent on the quality and diversity of the training data provided. 
Moving beyond conventional machine learning methods, neural networks have gained prominence in this domain \cite{guo2021robust}. For instance, convolutional neural networks (CNN) were employed by Kaliyar et al. \cite{27}, recurrent neural networks (RNN) were used by Jiang et al. \cite{28}, and Nasir et al. \cite{26} implemented a hybrid of CNN and RNN. Recently, researchers have explored pre-trained language models. BERT \cite{15} and RoBERTa \cite{25} are employed to analyze news content, achieving notable advancements. Nevertheless, due to their relatively modest model sizes, the capacity of these pre-trained models to extract complex knowledge and perform advanced reasoning is constrained, limiting their effectiveness in handling fake news that requires deeper content analysis and inference. In contrast, large language models (LLMs), such as GPT \cite{29}, exhibit superior performance in natural language processing (NLP) tasks by employing deeper neural architectures and significantly larger parameter counts. These models rely on extensive textual datasets from diverse fields and topics, constructing a comprehensive knowledge base and contextual understanding that bolsters their reasoning capabilities. Consequently, LLMs require minimal additional data for fine-tuning to effectively differentiate authentic news from misinformation \cite{30,31,32}. However, most existing studies have predominantly focused on either text or image data in isolation, rather than integrating both modalities for a more comprehensive approach.

\subsection{Multimodal Classification}
In the wake of the incessant development of social media, the news that is now being widely spread predominantly incorporates information such as text and images. As a result, scholars have stepped up their endeavors in the detection of multimodal fake news. Singhal et al. \cite{12} introduced a multimodal model which harnesses text and visual features. Likewise, Giachanou et al. \cite{13} combined the image features extracted by the VGG \cite{14} model and the text features extracted by the BERT \cite{15} model to detect image-text misinformation. Aneja et al. \cite{16} centered their attention on "Cheapfakes" produced by employing free artificial intelligence methods (e.g. filtering). They made use of multimodal embedding to predict whether image-caption pairs are mismatched. Fact verification, on the other hand, necessitates an additional information base. In light of this circumstance, some scholars have endeavored to utilize LLMs with pre-trained on the prior knowledge. Zhu et al. \cite{17} introduced the multimodal large language model MiniGPT-4, which achieves alignment between image and linguistic features by employing the Q-Former module. Liu et al. \cite{18} introduced a fake news detection model, FakeNewsGPT-4, leveraging the MiniGPT-4 framework. This model advances the prompting capabilities of large language models by integrating both prior and dynamically generated knowledge, thereby achieving superior performance across multiple domains.  %Qi et al. \cite{19} focused specifically on detecting off-topic fake news through large language models. They introduced the model SNIFFER, which uses prompts derived from news samples to enhance training on the detection model, thereby improving its effectiveness in identifying off-topic misinformation.

\section{Methodology}
\label{sec:3}
In this paper, we propose an innovative model for the task of multimodal fake news detection, as illustrated in Figure \ref{framework}. The overall structure comprises three core components: the contrastive module, the multimodal fusion module, and the classification module. To overcome the lack of training data, a contrastive learning mechanism is incorporated. In this mechanism, the image feature is acquired through augmentation where the model is trained by maximizing the similarity between positive pairs (e.g., different augmentations of the same data instance) and minimizing the similarity between negative pairs (from different data instances). Once the image features has been learned, we then use the pre-trained large-scale model to align the text content with the image features, which we called the infuse module. Rather than directly using the image features extracted from contrastive learning, we introduced a multi-task learning approach namely Q-Former to dynamically adjust the feature weights of image, thus to achieve the co-optimization of the image feature encoder and the text encoder. Subsequently, the pre-trained large-scale model, which is structured upon MiniGPT-4, is employed to effect a profound combination of image and text features. The multimodal features along with the appropriate prompts are fed as input of large language model. The output is followed by the linear layer and finally is trained to classify an accurate inference regarding the authenticity of news. We will explain each module in detail as follows:

\subsection{The Contrastive Learning Module}
Within the contrastive learning module, each of the image undergoes augmentation procedures. Subsequently, the augmented images are inputted into an image encoder, which is then followed by a fully connected layer. This sequential process is designed to learn the image feature and generate a dense vector. For the purpose of training the contrastive learning model, all the augmented images originating from the same sample are regarded as positive instances. Meanwhile, images that are randomly selected from other samples within the dataset are considered as negative instances. To be specific, the image is augmented (e.g. rotation, flip, scaling etc.) and then fed to the encoder for feature extraction. We adopt the pre-trained ViT model \cite{22} as the backbone network.

\begin{figure}[t]
\centering
\includegraphics[scale=0.6]{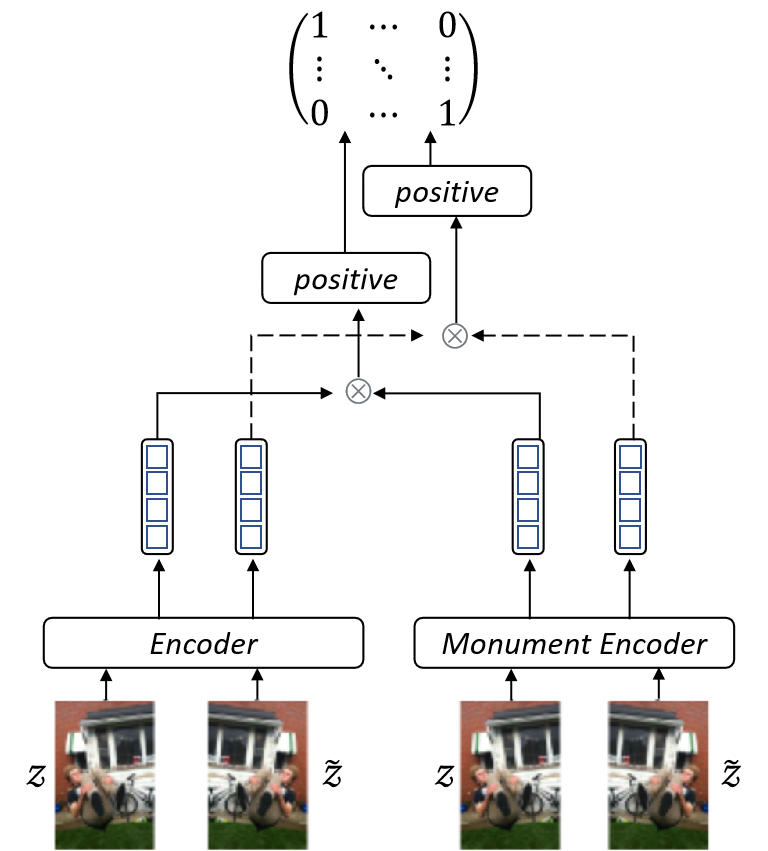}
\caption{Monument configuration for contrastive learning}
\label{momentum}
\end{figure}

Rather than directly training ViT for image feature extraction, we harness the power of momentum mechanism to smooth and stabilize the image encoding. As shown in Fig.~\ref{momentum}, the input image $Z$, along with the augmentation $\tilde{Z}$, are fed to the the image encoder and momentum encoder. Both of them are identical initially while performing training procedure individually. The difference is that the parameters of the momentum encoder are not updated synchronously with those of the encoder. Updating the momentum encoder parameters using a smaller value ensures the stability of the training process, as shown in Equation \ref{eq:moment}:
\begin{equation}
    \begin{aligned}
    y_{t}=m y_{t-1}+(1-m) x_{t}
    \end{aligned}
\label{eq:moment}
\end{equation}

where $y_{t-1}$ is the parameter at the previous moment, $x_t$ is the parameter at the current moment, and $m$ is the updated threshold of the parameter. The image features from two encoders are conducted scalar product respectively to get two matrices. Then, the two matrices are added to get final matrix for contrastive learning. We use mini-batch for training. The elements of the diagonal of the matrix indicate the similarity between the positive samples within a mini-batch, and the other elements indicate the similarity between the positive samples and the negative samples. The overall loss function of contrastive learning is using InfoNCE as shown in Equation \ref{eq:infonce}:
\begin{equation}
    \begin{aligned}
    L_{1}=-\log \frac{\exp \left(q \cdot k^{+} / \tau\right)}{\exp \left(q \cdot k^{+} / \tau\right)+\sum_{k^{-}} \exp \left(q \cdot k^{-} / \tau\right)}
    \end{aligned}
\label{eq:infonce}
\end{equation}
where $\tau$ is the temperature coefficient that controls the softness or sharpness of the probability distribution over the positive and negative samples when calculating the InfoNCE loss. $q$ is an anchor sample, $k^+$ is the positive sample, and $k^-$ is the negative sample. Mathematically, the InfoNCE loss is given by a formula that involves taking the logarithm of the ratio of the exponential of the score of the positive sample to the sum of the exponentials of the scores of all samples (both positive and negative).

\subsection{Multimodal Fusion Module}
Concurrently, with regard to the text content, the Byte-Pair Encoding algorithm is initially employed as a tokenizer for the purpose of transforming sentences into tokens. Subsequently, the multimodal model predicated on MiniGPT-4 amalgamates the text along with the image features that have been encoded during the pre-training phase. Given that the large language model (LLM) utilized in our paper has already undergone pre-training with an extensive volume of data in advance, the downstream task merely necessitates fine-tuning with a relatively small quantity of data to fulfill the specified task requirements.

\begin{figure}[t]
\centering
\includegraphics[scale=0.52]{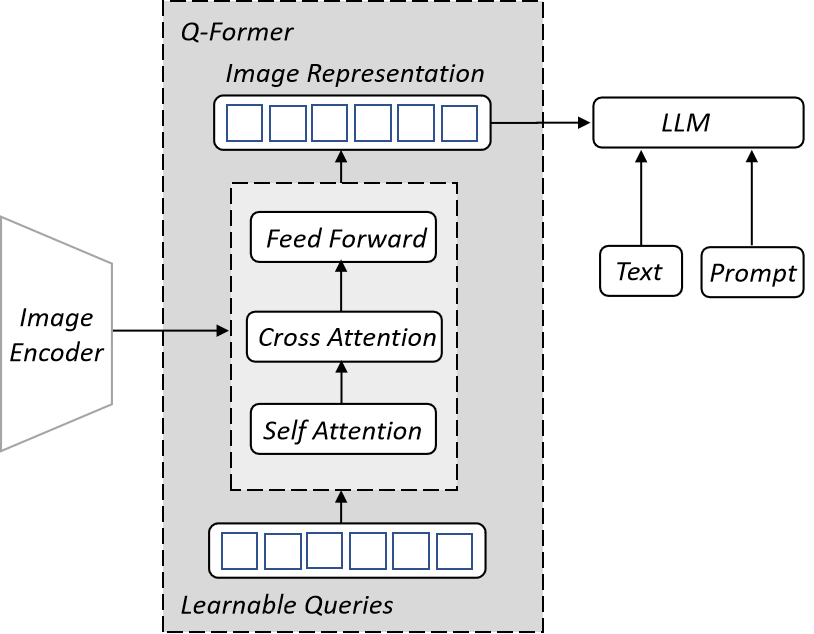}
\caption{Q-Former structure adopted from \cite{24}}
\label{qformer}
\end{figure}

Instead of using the image features from contrastive learning module, we would like to leverage Q-Former \cite{24}, a core component of MiniGPT-4 \cite{8}, to bridge the relationship between image and text. It extracts features from a frozen image encoder and aligns them with a large language model. The key component is a learnable query which is a set of vectors that are designed to interact with the input data in a way that helps extract relevant information and establish meaningful connections. The structure is illustrated in Fig.~\ref{qformer}. Q-Former's alignment is divided into two phases, the first phase mainly involves the learning of image features, through which Q-Former learns the most relevant image feature representation to the current text by the learnable queries. The second stage is generative learning, which combines the output of Q-Former with a pre-trained frozen large language model to achieve visual-to-language generative learning. The LLM (Vicuna \cite{33}) is used to understand and describe the visual expression features of the Q-Former output, thus building a relationship between visual information and linguistic description.

In the end of multimodal fusion module, text feature $e_{text}$ is obtained through the text embedding layer, image feature $e_{img}$ is obtained in a pre-trained image encoder, and function $f$ that can be interpreted by the large language model are obtained by using Q-Former with approapriate prompt, and these features are connected together to obtain the final mixed feature representation $E$. The equations are shown in Equation \ref{eq:eprompt} and \ref{eq:efeature}.
\begin{equation}
    \begin{aligned}
    e_{\text{prompt}}=e_{\text{text}}+f_{Q-\text{Former}}\left(e_{\text{img}}\right)
    \end{aligned}
\label{eq:eprompt}
\end{equation}

\begin{equation}
    \begin{aligned}
    E=LLM\left(e_{\text{prompt}}\right)
    \end{aligned}
\label{eq:efeature}
\end{equation}

\subsection{Classification Module}

After performing multimodal fusion module to obtain the hidden layer features $E$, these features are then input into the fake news classifier. Subsequently, the final authenticity of the news is output. The fake news classifier is composed of two linear layers along with a GELU activation function. The specific details of the classification process are presented in Equations \ref{eq:classifier} below:
\begin{equation}
    \begin{aligned}
    L_{2}=M L P(E \mid \theta)
    \end{aligned}
\label{eq:classifier}
\end{equation}
where $\theta$ is all the model parameters and $l$ is the specific news category which is true or false.

\subsection{Prompt}
Within the LLM domain, prompts hold a vital position as they significantly influence the model's actions and resultant outputs. Essentially, prompts are text-based instructions or input indications furnished to the LLM, with the aim of communicating the task or context that the model is required to manage. In our task, we crafted a variety of prompts allow the model be able to differentiate the authentic of news from different perspectives. The methodology put forward within this chapter employs four distinct Prompts for the training of the model. These Prompts are designed in line with the session format of the Vicuna language model and are randomly chosen to be inputted into the model. The specific details are presented as follows:
\begin{itemize}
\item \textit{\textless Img\textgreater\textless ImageHere\textgreater\textless /Img\textgreater\textless  Text\textgreater \\ \textless TextHere\textgreater\textless/Text\textgreater}Determine if the text and images of this news story are lying? \#\#\#Assistant:

\item \textit{\textless Img\textgreater\textless ImageHere\textgreater\textless /Img\textgreater\textless Text\textgreater \\ \textless TextHere\textgreater\textless/Text\textgreater}
Determine if this text and images are describing a rumor? \#\#\#Assistant:

\item \textit{\textless Img\textgreater\textless ImageHere\textgreater\textless /Img\textgreater\textless Text\textgreater \\ \textless TextHere\textgreater\textless/Text\textgreater}
Determine if this news story is false? \#\#\#Assistant:

\item \textit{\textless Img\textgreater\textless ImageHere\textgreater\textless /Img\textgreater\textless Text\textgreater \\ \textless TextHere\textgreater\textless/Text\textgreater}
Determine if this text and image information is untrue? \#\#\#Assistant:
\end{itemize}
In these Prompts, \textit{\textless ImageHere\textgreater} represents the visual features generated by the image encoder, while \textit{\textless TextHere\textgreater} represents the corresponding news text content.

\subsection{Loss Function}
In the model previously described, both the contrastive learning and classification components necessitate training procedures. As a result, the overall loss function is constituted by two sub-functions, namely the contrastive learning loss $L_{1}$ and the classification loss $L_{2}$. Given the varying optimization priorities between these two individual tasks, employing fixed weights does not yield optimal results, and manually adjusting these weights is time-consuming. To address this issue, we implemented the Automatic Weighted Loss (AWL) method \cite{34} to augment multi-task learning and concurrently optimize multiple loss functions. The details of this implementation are illustrated in Equation ~\ref{eq:multil}.

\begin{equation}
    L \approx \frac{1}{2 \sigma_{1}^{2}} L_{1}+\frac{1}{\sigma_{2}^{2}} L_{2}+\log \left(1+\sigma_{1}\right)+\\
    \log \left(1+\sigma_{2}\right)
\label{eq:multil}
\end{equation}

Where $\sigma_{1}$ and $\sigma_{2}$ are learnable parameters representing the uncertainty of the corresponding task, the higher the uncertainty of the task, the smaller the weight of its loss function. Through this dynamic adjustment, the model can learn the weights suitable for different tasks and avoid the negative impact of fine-tune of pre-trained model. Additionally, during the preliminary experiment, we found that the loss function was getting smaller, but the model was not improved. To avoid this issue, we add 1 to each of learnable parameters.

\section{Experimental Settings}
\label{sec:4}
\subsection{Datasets}
The data \cite{8} obtained from social media platforms generally comprises a variety of elements. Among them, invalid information such as URLs and stop words are frequently encountered. This study aims to remove such information. However, certain proper names, including those of individuals, locations, and countries, which are considered as key information, are deliberately retained on account of the implementation of LLMs. For the image data, a series of augmentation operations are meticulously devised. These encompass operations such as horizontal flipping, hue transformation, and grayscale conversion. During the training phase of the model, one of these augmentation operations is randomly selected and applied. In addition, the data associated with entries lacking attached images and those accompanied by invalid images are removed from the dataset. Following this pre-processing, a complete and refined dataset is constructed. This dataset comprises nearly 563,600 training samples, 59,000 validation samples, and 59,500 testing samples. The statistical details of the dataset are presented in Table~\ref{tab:dataset} below.

\begin{table}[htbp]
\caption{Statistical Information of the Dataset}
\label{tab:dataset}
\centering
\scalebox{1.2}{
\begin{tabular}{ccc}
\toprule
Data Category & True & False \\
\midrule
Training Set & 222.1k & 341.5k \\
Validation Set & 23k & 36k \\
Testing Set & 23.5k & 36k \\
\bottomrule
\end{tabular}}
\end{table}

\subsection{Implementation Details}
The image encoder of our network employed off-the-shelf ViT model where its block size is set to 14. All input images were scaled to a fixed resolution of 224 × 224. Followed by the encoding process, the feature vector dimension was set up to the batch size × 12 × 1048. In terms of large language model we used, it is worth noting that the Vicuna model applied in this paper is version 7B. During the training process, a total of 100 training epochs were conducted, which we proved it's the best practices based on a variety of preliminary experiments, we extended the number of samples in each batch to 96 by the gradient accumulation method. The assessment methods in this paper are chosen as the accuracy, precision, recall, and F1-score, which are widely used in the field of fake news detection, as the metrics for evaluating the performance of the model.

\subsection{Baselines}
To assess the efficacy of our proposed model, this study selects a series of state-of-the-art classification methods as the baselines for comparing. All the chosen models are widely employed in image-text paring tasks and achieved strong ability to detect fake news. All models were trained using identical publicly available datasets and tested on the same data. Various evaluation metrics were carried out. Furthermore, we investigated model performance under conditions of limited training data, specifically utilizing only 10\% of the Fakeddit training set. The baseline methods are: \textbf{EANN (2018)} \cite{1} employs the pre-trained VGG for extracting image features, followed by Text-CNN. EANN incorporates auxiliary tasks such as an event discriminator, which outputs event categories of news to aid in decision-making. \textbf{CAFE (2022)}\cite{2} leverages pre-trained BERT and ResNet-34 models to extract features from text and image respectively. It used a unified embedding space in order to integrate diverse modalities. \textbf{SpotFake (2019)} \cite{3} utilizes a pre-trained VGG network to extract features from images and BERT to capture textual features. SpotFake's notable advantage lies in its simplicity, avoiding complex auxiliary training tasks often seen in other models. \textbf{SpotFake+(2019)} \cite{4} modified SpotFake framework by the use of additional layers with attention mechanisms, advanced regularization techniques, and more sophisticated training process. \textbf{MVAE (2019)} \cite{5} The Multimodal Variational Autoencoder (MVAE) extends variational autoencoders for multimodal data (text, images, audio). It learns shared latent representations to address tasks like classification. \textbf{HMCAN (2021)} \cite{6} is short for Hierarchical Memory Compressed Attention Network (HMCAN) which introduces a hierarchical structure with memory-compressed attention to capture both local and global contexts, optimizing long-sequence processing in NLP tasks. \textbf{VERITE (2024)} \cite{7}: “VERification of Image-TExt pairs” (VERITE) uses CLIP (Contrastive Language–Image Pre-training) as the feature extractor for images and texts, followed by the encoding layer of the Transformer. It has demonstrated a strong fake news detection performance.

\begin{table}[h]
\centering
\caption{The results of three models over the Accuracy, Precision, Recall and F1-score.}
\label{result_tab}
\scalebox{1.0}{
\begin{tabular}{ccccc}
\toprule
\textbf{Model} & \textit{Accuracy} & \textit{Precision} & \textit{Recall} & \textit{F1-Score} \\
\midrule
\textit{\textbf{EANN}} \cite{1} & 72.27 & 78.43 & 63.4 & 70.12 \\
\textit{\textbf{CAFE}} \cite{2} & 84.14 & 85.39 & 85.27 & 85.32 \\
\textit{\textbf{SpotFake}} \cite{3} & 77.29 & 71.63 & 70.77 & 71.20 \\
\textit{\textbf{SpotFake+}} \cite{4} & 83.08 & 86.38 & 84.87 & 85.62 \\
\textit{\textbf{MVAE}} \cite{5} & 70.24 & 76.53 & 74.75 & 75.63 \\
\textit{\textbf{HMCAN}} \cite{6} & 82.89 & 84.03 & 84.04 & 84.03 \\
\textit{\textbf{VERITE}} \cite{7} & 84.72 & 85.34 & 84.37 & 84.85 \\
\textit{\textbf{Ours}} & \textbf{88.88} & \textbf{86.40} & \textbf{85.40} & \textbf{85.90} \\
\bottomrule
\end{tabular}}
\end{table}

\section{Results}
\subsection{General Performance}
The Table~\ref{result_tab} presents a comparative analysis of various models based on four evaluation metrics: accuracy, precision, recall, and F1-score. The models compared include EANN, CAFE, SpotFake, SpotFake+, MVAE, HMCAN, VERITE, and the proposed model referred to as "Ours." EANN achieved the lowest accuracy of 72.27\% and an F1-score of 70.12\%. In contrast, CAFE performed significantly better, with an accuracy of 84.14\% and an F1-score of 85.32\%, indicating robust performance in both precision and recall. Comparing SpotFake and its enhanced version, SpotFake+ showed notable improvement with SpotFake+ reaching an accuracy of 83.08\% and an F1-score of 85.62\%, outperforming SpotFake's 77.29\% accuracy and 71.20\% F1-score. HMCAN demonstrated superior precision, recall, and F1-score values, all around 84\%, compared to MVAE's lower performance in these metrics. VERITE model also showed high performance with an accuracy of 84.72\% and an F1-score of 84.85\%, closely aligning with CAFE and SpotFake+ in terms of overall effectiveness.

The proposed model ("Ours") outperformed all baseline models with the highest accuracy of 88.88\%, precision of 86.40\%, recall of 85.40\%, and F1-score of 85.90\%. This indicates the superior capability of the our method in handling the experimental tasks, achieving a balanced and high performance across all evaluation metrics. The observed improvements in accuracy and recall of our model indicate a significant reduction in the likelihood of misclassifying real news articles as fake, as well as a substantial decrease in the incidence of false positives. 
%The remarkable performance exhibited by the proposed approach can be attributed to two main factors. Firstly, the extensive pre-training procedure conducted on a large-scale dataset has enabled the model to possess a profound understanding of the nuances in the language along with visual cues. Secondly, the employment of a more substantial model compared to alternative approaches further enhanced its capabilities. Collectively, these factors boost the model's reasoning ability, facilitating a more refined integration and comprehension of both textual and visual information.

To further analysing results, EANN and MVAE exhibit only slight performance differences, reflecting the limitations of their basic multimodal fusion approaches. EANN utilizes an event discriminator for classification but does not address the semantic relationships between text and image features. MVAE, while using decoding structures to improve performance, still struggles with misalignment between modalities. SpotFake outperforms both by using pre-trained BERT for text encoding, although the direct concatenation of features limits its potential. SpotFake+ achieves even higher accuracy with XLNet, though it faces the same fusion constraints. HMCAN, CAFE, and VERITE outperform others, but fall short compared to the proposed method. HMCAN, about 6\% less accurate than ours, applies multimodal contextual attention to fuse features. CAFE improves accuracy by aligning features in a unified space, reducing semantic gaps. VERITE stands out by combining CLIP for feature extraction with attention-based fusion, making it the top performer among the comparative models.

Based on the above experimental results, this paper believes that the main factors that affect classification performance are as follows: 
\begin{itemize}
    \item The approach of feature fusion exerts a substantial influence on the performance of the model. While the direct concatenation of features appears to be a straightforward method, it fails to facilitate the interaction among data from diverse modalities. In contrast, models engineered with feature fusion techniques tend to attain favorable accuracy levels.
    \item Model scale also serves as a crucial metric. Apart from the model introduced in this study, the parameter scales of other comparative models are relatively limited. This smaller parameter size constrains the model's inferential abilities, impeding its capacity to make precise judgments regarding text and image features.
    %\item This method has undergone extensive data pre-training, enabling it to capture complex features and patterns inherent in multimodal data.
\end{itemize}

\subsection{Ablation Study}
In this section, a series of ablation experiments were carried out. These experiments strictly adhered to the principle of controlling variables, ensuring that only one module was altered each time, thereby enabling the individual assessment of the impact of each module on the overall performance. Specifically, two types of experiments were designed, namely the comparison of different modal data and comparison of individual module. The specific settings of the experiments are as follows:
\subsubsection{Comparison of Different Modal Data}
To verify the impact of different modal data on fake news detection, three experiments were designed respectively, namely single-modal image, single-modal text, and the multimodal combination of image and text. According to the experimental results presented in Table~\ref{result_modals}, in the experiment where only single-modal image data was utilized, the lowest accuracy, precision, and F1-score were 79.68\%, 68.76\%, and 72.75\% respectively. In contrast, the model using single-modal text data achieved relatively higher accuracy, precision, and F1-score, which were 83.18\%, 79.71\%, and 78.64\% respectively. This indicates that text information is of greater significance in determining the authenticity of news compared to image information. The data of the multimodal combination of image and text yielded the highest accuracy of 88.88\%, which is approximately 7\% and 5\% higher than that of single-modal image and text respectively. Moreover, the accuracy, recall rate, and F1-score were all at relatively high levels, thereby demonstrating the effectiveness of multimodal fusion in fake news detection.

\begin{table}[h]
\caption{Ablation Comparison of Different Modals}
\centering
\scalebox{1.1}{
\begin{tabular}{cccccc}
\toprule
Modal Type & Accuracy & Precision & Recall & F1-score \\
\midrule
Image & 79.68 & 68.76 & 89.38 & 72.75 \\
Text & 83.18 & 79.71 & 77.25 & 78.46 \\
Image and Text & 88.88 & 86.40 & 85.40 & 85.90 \\
\bottomrule
\end{tabular}}
\label{result_modals}
\end{table}

\subsubsection{Comparison of Individual Modules}
In order to precisely gauge the contributions of each module within the proposed methodology to the performance of the model, three sets of comparative experiments were meticulously designed for this ablation study. The detailed experimental configurations are presented as follows:
\begin{itemize}
    \item Experiment A: Employed the large language model along with the fully connected layer.
    \item Experiment B: Incorporated the contrastive learning module on top of the setup in Experiment A.
    \item Experiment C: Added the multimodal infuse module based on the configuration of Experiment B.
\end{itemize}
As shown in Table~\ref{tab:ablation_comparison}, Experiment A displayed the weakest performance in the context of fake news detection. Its accuracy rate was merely 87.16\%, which was lower than that attained by Experiment B and Experiment C. The models that utilized contrastive learning and multimodal learning approaches demonstrated an advantage over the LLM model across various metrics. The improvement could be ascribed to the data augmentation procedure, which broadened the data samples and thus improved the generalization and robustness of the model. When analyzing the precision and recall rates, it was found that the multi-task learning method maintained a relatively stable performance, while Experiments A and B showed relatively higher recall rates.% This difference arose from the fact that the multi-task learning method balances the learning rates of different tasks, thereby preventing the overemphasis on a single task from hindering the learning of other tasks. This experiment further substantiates the efficacy of each module expounded within this section.

\begin{table}[htbp]
\centering
\caption{Ablation Comparison of Different Modules}
\label{tab:ablation_comparison}
\scalebox{1.1}{
\begin{tabular}{ccccc}
\toprule
Model & Accuracy & Precision & Recall & F1-score \\
\midrule
Experiment A & 87.16 & 78.60 & 92.91 & 85.16 \\
Experiment B & 88.21 & 81.19 & 91.45 & 86.02 \\
Experiment C & 88.88 & 86.40 & 85.40 & 85.90 \\
\bottomrule
\end{tabular}}
\end{table}

\section{Conclusion}
\label{sec:5}
This study aimed to establish an innovative multimodal classification framework for verifying the authenticity of news shared on social media platforms. In light of limited labeled data, especially for image-based content, we employed contrastive learning to improve feature representation. Additionally, we demonstrated the effectiveness of Large Language Models (LLMs) in facilitating the seamless integration of text and image features. Instead of a simplistic multimodal fusion, we introduced a learnable alignment module that significantly improved the model's accuracy by aligning text-image features. Key contributions of this work include: (1) We developed an enhanced fake news detection model grounded in contrastive learning, a self-supervised approach utilizing data augmentation during training. Experimental results strongly indicated the model’s superiority in scenarios with limited labeled data. (2) Our focus on news items as paired image-text combinations revealed that dynamically infusing features from different data formats substantially improved fake news detection, achieving an accuracy rate close to 89\%. This multimodal approach considerably outperformed single-modal text- or image-only analyses. (3) By integrating contrastive learning with LLMs through a carefully designed feature infusion mechanism for multimodal classification, we conducted extensive comparative experiments. The findings highlighted the robust detection capabilities of our proposed model relative to numerous existing models, and demonstrated that larger LLMs further enhance detection accuracy. However, a primary limitation of this study lies in its focus solely on the data itself, excluding the potential value of supplementary information such as social networks, geographic data, and event context. Future research will aim to enhance integration strategies between aforementioned model with such content, concentrating on methods to further improve the accuracy of misinformation detection on social media platforms.
\bibliographystyle{plain}
\bibliography{ref}

\end{document}